\documentclass[journal]{IEEEtran}
\IEEEoverridecommandlockouts
\usepackage{amsmath,amssymb,amsfonts}
\usepackage{algorithmic}
\usepackage{graphicx}
\usepackage{textcomp}
\usepackage{csquotes}
\usepackage{xcolor,soul}
\usepackage{adjustbox}
\usepackage{hyperref}
\usepackage{multirow}
\usepackage{subcaption}
\usepackage{makecell}

\usepackage[numbers,sort&compress]{natbib}

\usepackage{etoolbox}

\usepackage{amsmath}

\usepackage{mathtools}
\usepackage{gensymb}
\usepackage{multirow}
\usepackage{enumitem}
\usepackage{diagbox}
\usepackage{todonotes}
\usepackage{parskip}

\makeatletter
\newcommand{\thickhline}{%
    \noalign {\ifnum 0=`}\fi \hrule height 1.5pt
    \futurelet \reserved@a \@xhline
}

\begin{document}
\title{DCNet: A Data-Driven Framework for DVL Calibration}

\author{\IEEEauthorblockN{Zeev Yampolsky\IEEEauthorrefmark{1} and Itzik Klein}
\IEEEauthorblockA{\\The Hatter Department of Marine Technologies\\
Charney School of Marine Sciences, University of Haifa,
Haifa, Israel}

\thanks{\IEEEauthorrefmark{1}Corresponding author: Zeev Yampolsky, zyampols@campus.haifa.ac.il.}}

\maketitle
\begin{abstract}
Autonomous underwater vehicles (AUVs) are underwater robotic platforms used in a variety of applications. An AUV's navigation solution relies heavily on the fusion of inertial sensors and Doppler velocity logs (DVL), where the latter delivers accurate velocity updates. To ensure accurate navigation, a DVL calibration is undertaken before the mission begins to estimate its error terms. During calibration, the AUV follows a complex trajectory and employs nonlinear estimation filters to estimate error terms. In this paper, we introduce DCNet, a data-driven framework that utilizes a two-dimensional convolution kernel in an innovative way. Using DCNet and our proposed DVL error model, we offer a rapid calibration procedure. This can be applied to a trajectory with a nearly constant velocity. To train and test our proposed approach a dataset of 276 minutes long with real DVL recorded measurements was used. We demonstrated an average improvement of 70\% in accuracy and 80\% improvement in calibration time, compared to the baseline approach, with a low-performance DVL. As a result of those improvements, an AUV employing a low-cost DVL, can achieve higher accuracy, shorter calibration time, and apply a simple nearly constant velocity calibration trajectory. Our results also open up new applications for marine robotics utilizing low-cost, high-accurate DVLs.
\end{abstract}

\section{Introduction}\label{intro_sec}
\noindent Autonomous underwater vehicles (AUVs) play a significant role in a variety of marine and underwater applications. For example, oceanographic surveys \cite{paull2013auv, wynn2014autonomous}, underwater structure inspection such as gas pipelines \cite{jacobi2015autonomous,zhang2022submarine,zhang2021subsea}, and detection of oil leaks \cite{li2018autonomous,vasilijevic2015auv}. During the AUVs mission, the navigation solution is critical for the mission's success \cite{leonard2016autonomous,xu2021accurate}. The navigation solution, which is the position, velocity, and orientation of the AUV, is commonly provided by inertial sensors, Doppler velocity log (DVL), acoustic positioning, and other sensors such as pressure sensors \cite{mu2021practical,ding2023rd}.\\
The DVL is an acoustic sensor that utilizes the Doppler frequency shift effect. It transmits acoustic beams to the seafloor, which are then reflected back. By using the frequency shift, the DVL is able to calculate each beam's velocity and then to estimate the AUV velocity \cite{10674766,brokloff1994matrix}.\\
The DVL measurements are subject to different types of error sources and noise that influence the measurement quality and accuracy \cite{levy2023ins,cohen2023set}. To cope with such effects, DVL calibration is performed prior to mission start. In such a process, the error terms are estimated and calibrated for \cite{xu2022novel,wang2021quasi}.
Commonly, the calibration process requires the AUV to sail at sea level while using a reference velocity sensor to determine the AUV's ground truth (GT) velocity. Commonly, a global navigation satellite system (GNSS) receiver with real-time kinematics (GNSS-RTK) is used as a reference sensor since it provides centimeter-level accuracy \cite{li2018high}. \\
During the calibration procedure, the AUV is required to follow long and complex trajectories, which in turn employ a nonlinear estimation filter, resulting in an overall complex calibration process \cite{ning2023research}. Typically, the DVL calibration algorithms are model-based approaches employing estimation filters such as the Kalman filter, adaptive Kalman filter \cite{liu2018innovative,xu2020novel}, and the invariant extended Kalman filter \cite{xu2022novel}. Other approaches propose employing genetic algorithms for the DVL calibration \cite{liu2022simultaneous}. \\
Data-driven methods have been employed in AUV-related tasks with promising outcomes. For example, deep-learning frameworks have been used to estimated missing DVL beams in partial and complete outage scenarios \cite{cohen2023set,yona2024missbeamnet,yao2022virtual}. Additionally, when all beams are available, the BeamsNet approach offers a more accurate and robust velocity solution using a dedicated deep-learning framework \cite{cohen2022beamsnet}.\\
Recently, we proposed a data-driven method for DVL calibration \cite{yampolsky2024dvl}. There, a procedure that employs a simple constant velocity trajectory and a data-driven framework has demonstrated, on simulative data, effective calibration performance.
In this research we leverage our initial work and propose a simple, yet effective, DVL calibration process. To meet this goal we make the following contributions:
\begin{enumerate}
    \item \textbf{DVL error models}: Five error models are proposed and investigated to relate the measured DVL velocity to the reference GNSS-RTK velocity. Each proposed error model utilizes a different combination of the DVL error terms including scale factor and bias.
    \item \textbf{DCNet}: A simple, yet efficient, end-to-end regression network for the DVL calibration process is developed. DCNet is a convolution based framework that employs a two-dimensional dilated convolution kernel designed to process the input data in a unique manner.
    \item \textbf{Data-driven calibration}: We propose a data-driven framework for the calibration process. Utilizing, our proposed error models and DCNet, our framework can be applied on a simple nearly constant velocity trajectory and  requires less time for the calibration process.
\end{enumerate}
\noindent To demonstrate our proposed approach, we employed a dataset consisting of $276$ minutes. The data was recorded using the University of Haifa's AUV, the "Snapir" \cite{cohen2022beamsnet,cohen2024kit}, in two separate experiments totaling in ten recorded different trajectories. 
Our proposed data-driven approach achieved an average of $70\%$ accuracy improvement, and a $80\%$ improvement in calibration time, over the baseline approach on a low-grade DVL, using real-world recorded data. As a result of our work, low-cost DVLs can be used to achieve higher accuracy and reduce costs. This will open up new applications for marine robotic systems using low-cost DVLs with high accuracy.\\
The rest of this paper is organized as follows: Section \ref{prob_form_sec} describes the DVL velocity measurement and the baseline calibration approach. Section \ref{prop_approach} presents our proposed data-driven approach. Section \ref{res_sec} gives the experimental results and Section \ref{conc_sec} gives the conclusions of this research.

\section{Problem Formulation}\label{prob_form_sec}
\subsection{DVL Velocity}\label{dvl_basic_eq}
\noindent For a DVL with an "$\times$" configuration, referred to as the "Janus" configuration, the direction of each beam in the DVL's body frame can be expressed as \cite{cohen2022beamsnet,braginsky2020correction}:
\begin{equation}\label{bi_in_h}
    \centering
        \boldsymbol{b}_{\dot{\imath}}=
        \begin{bmatrix} 
        \cos{\psi_{\dot{\imath}}}\sin{\alpha}\quad
        \sin{\psi_{\dot{\imath}}}\sin{\alpha}\quad
        \cos{\alpha}
    \end{bmatrix}_{1\times3}
\end{equation}
where $\psi_{i}$ and $\alpha$ refer to the pitch and yaw angles of beam $i = 1,2,3,4$, respectively. For all beams, the pitch angle remains constant, whereas the yaw angle is given by
\begin{equation}\label{yaw_of_beams}
    \centering
        \psi_{\dot{\imath}}=(\dot{\imath}-1)\cdot\frac{\pi}{2}+\frac{\pi}{4}\;[rad]\;,\; \dot{\imath}=1,2,3,4.
\end{equation}
By stacking all beam projection vectors from \eqref{bi_in_h}, the transformation matrix $\mathbf{H} \in \mathbb{R}^{4 \times 3}$ is formed:
\begin{equation}\label{H_mat_defenition}
    \centering
        \mathbf{H}=
        \begin{bmatrix} \boldsymbol{b}_{1}\\\boldsymbol{b}_{2}\\\boldsymbol{b}_{3}\\\boldsymbol{b}_{4}\\
    \end{bmatrix}_{4\times3}
\end{equation}
The estimated velocity vector, expressed in the DVL frame, is
\begin{equation}\label{psudo_inverse}
    \centering
    \boldsymbol{v}_{\mathrm{AUV}}^{d} = (\mathbf{H}^{T}\mathbf{H})^{-1}\mathbf{H}^{T}\boldsymbol{y}
\end{equation}
The DVL error model is given by \cite{10674766}:
\begin{equation}\label{y_as_func_of_H_and_v_error_model}
    \centering
        \tilde{\boldsymbol{y}} = [\mathbf{H} \boldsymbol{v}_{\mathrm{AUV}}^{d}(1+\boldsymbol{s}_{\mathrm{DVL}})]+\boldsymbol{b}_{\mathrm{DVL}}+\boldsymbol{\sigma}_{\mathrm{DVL}}
\end{equation}
\noindent where $\tilde{\boldsymbol{y}}$ is the measured beam velocity vector, $\boldsymbol{v}_{\mathrm{AUV}}^{d}$ is the AUV velocity expressed in the DVL frame, $\boldsymbol{s}_{\mathrm{DVL}}$ is the scale factor applied over the beams, $\boldsymbol{b}_{\mathrm{DVL}}$ is the additive bias vector, and $\boldsymbol{\sigma}_{\mathrm{DVL}}$ is the additive zero mean Gaussian white noise vector.

\subsection{GNSS Based DVL Calibration}\label{gnss_dvl_relation}
\noindent During the calibration process the AUV follows a predetermined trajectory while sailing at sea level, allowing GNSS-RTK velocity measurements of the AUV, while in parallel maintaining sufficient depth for the DVL to provide measurements. The GNSS-RTK and DVL measurements are related using the following error model \cite{xu2022novel}:
\begin{equation}\label{dvl_to_gnss_eq}
    \centering
    \hat{\boldsymbol{v}}^{d} = (1+k)\mathbf{R}_{b}^{d}(\mathbf{R}_{n}^{b}
    \boldsymbol{v}^{n} + \boldsymbol{\omega}_{nb}^{b} \times \boldsymbol{l}_{\mathrm{DVL}}) + \boldsymbol{\delta v}^{d} 
\end{equation}
where $\boldsymbol{v}^{n}$ is the AUV reference velocity provided by the GNSS-RTK in the navigation frame, $\mathbf{R}_{n}^{b}$ is the transformation matrix from the navigation to the body frame, $\boldsymbol{\omega}_{nb}^{b}$ is the angular velocity of the AUV, and $\boldsymbol{l}_{\mathrm{DVL}}$ is the lever-arm between the DVL and the AUV center of mass. Matrix $\mathbf{R}_{b}^{d}$ is the constant and known transformation matrix from the DVL to the body frame, $k$ is the scale factor applied over the reference velocity in the DVL frame, and $\boldsymbol{\delta v}^{d}$ is the additive zero mean Gaussian white noise. Note that the term $\boldsymbol{\omega}_{nb}^{b} \times \boldsymbol{l}_{\mathrm{DVL}}$ is usually neglected since the lever-arm is generally small and can be measured and compensated \cite{xu2020novel}. Thus, \eqref{dvl_to_gnss_eq} reduces to:
\begin{equation}\label{gnss_dvl_fin_error_model}
    \centering
    \hat{\boldsymbol{v}}^{d} = (1+k){\mathbf{R}}_{n}^{d}
    \boldsymbol{v}^{n} + \boldsymbol{\delta v}^{d} 
\end{equation}
where $\mathbf{R}_{n}^{d}$ is the transformation matrix from the navigation from to the DVL frame.
By applying a vector norm on the error model presented in \eqref{gnss_dvl_fin_error_model}, the following is achieved \cite{liu2022gnss}:
\begin{equation}\label{scale_norm_eq_first}
    \mid\mid \hat{\boldsymbol{v}^{d}} \mid\mid = (1+k)\mid\mid {\hat{\mathbf{R}_{n}^{d}}}\boldsymbol{v}^{n} \mid\mid.
\end{equation}
Further development of \eqref{scale_norm_eq_first} produces the following scalar scale factor estimation:
\begin{equation}\label{scale_integ_continues}
    \hat{k}_{t} = \frac{ \mid\mid \hat{\boldsymbol{v}}^{d}_{t} \mid\mid}
    { \mid\mid \hat{\mathbf{R}_{n}^{d}}
    \boldsymbol{v}^{d}_{t} \mid\mid } - 1 , \ t = 1,\ldots,T
\end{equation}
where $\hat{k}_{t}$ is the estimated scalar scale factor at time step $t$ that is applied to the measured DVL velocity as presented in \eqref{gnss_dvl_fin_error_model}. Note that $\hat{k}_{t}$ is 1) a scalar scaling factor, thus the same scaling factor is applied over all three velocity axes, and 2) is an estimate at a single time step $t$, out of the entire calibration recording with $T$ time steps. To estimate a single scaling factor and mitigate the zero mean Gaussian white noise effect, the average of all $T$ time steps of $\hat{k}_{t}$ is calculated as follows: 
\begin{equation} \label{average_direct_scale}
    \centering
    \overline{k} = \frac{1}{T} \sum_{t=1}^{T} \hat{k}_{t}.
\end{equation}
This is the baseline approach, which provides the estimated scalar scale factor at the end of the calibration procedure.

\section{Proposed Approach}\label{prop_approach}
\noindent In this section we present five different DVL error models and the derivation of our DCNet approach.
\subsection{DVL Error Models}\label{prop_err_model}
\begin{figure}[!ht]
	\centering
		\includegraphics[width = 0.75\linewidth]{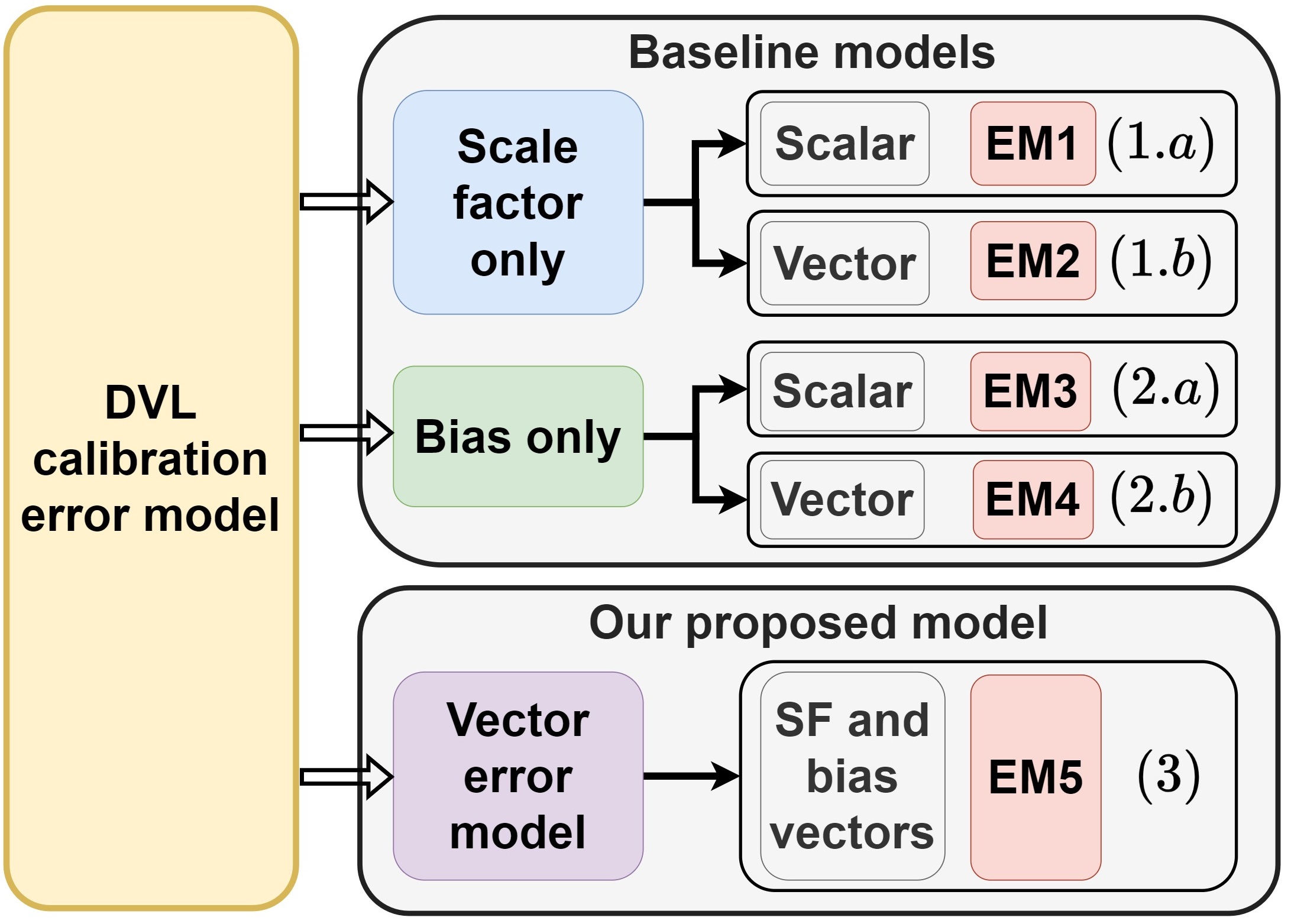}
	  \caption{Five DVL error models divided into baseline models \textbf{EM1} - \textbf{EM4}, and our proposed \textbf{EM5}.}\label{error_models_fig}
\end{figure}

\noindent In our preliminary work \cite{yampolsky2024dvl} we proposed the following DVL error model:
\begin{equation}\label{prop_general_error_model}
    \centering
    \hat{\boldsymbol{v}}^{d} = (1+\boldsymbol{k}_{\mathrm{DVL}})\hat{\mathbf{R}}_{b}^{d}
    \boldsymbol{v}^{n} + \boldsymbol{b}_{\mathrm{DVL}} + \boldsymbol{\delta v}^{d} 
\end{equation}
where $\boldsymbol{k}_{\mathrm{DVL}}$ is the scale factor vector:
\begin{equation}\label{scale_vec_def}
    \centering
    \boldsymbol{k}_{\mathrm{DVL}} = [k_{{x}} \ k_{y} \ k_{z}]^{T} \in \mathbb{R}^{3}
\end{equation}
 $\boldsymbol{b}_{\mathrm{DVL}}$ is the bias vector:
\begin{equation}\label{bias_vec_def}
    \centering
    \boldsymbol{b}_{\mathrm{DVL}} = [b_{x} \ b_{y} \ b_{z}]^{T} \in \mathbb{R}^{3}
\end{equation}
and $\boldsymbol{\delta v}^{d}$ is the zero mean Gaussian white noise.
Note that, as presented in \eqref{scale_vec_def} and \eqref{bias_vec_def}, the scale factor and bias error terms can be modeled as scalars or vectors. When only a single type of error term is presented (bias or scale factor), four different DVL error models can be deduced from \eqref{prop_general_error_model}, as we presented in \cite{yampolsky2024dvl}.\\
\noindent With a goal to estimate the velocity error in each axis, we propose a comprehensive error model including a scale factor vector and a bias vector.
Thus, in total, we consider five different DVL error models (EMs):
\begin{enumerate}\label{error_model_variations}
        \item \textbf{Scale Factor Only:} This model corrects DVL measurements using only a scale factor, as commonly addressed in the literature. Two alternatives are considered for the scale factor-only model:
            \begin{enumerate}
                \item \textbf{EM1 - Scalar Scale-Factor}: The scale factor has the same value $k_{x} = k_{y} = k_{z} = k$ in all three axes, and thus is addressed as a scalar.
                \item \textbf{EM2 - Vector Scale-Factor}: The scale factor values may differ in each axis as defined in \eqref{scale_vec_def}.
            \end{enumerate}
        \item \textbf{Bias Only:} The bias error model corrects DVL measurements using only a bias. Two alternatives are considered for the bias-only error model:
            \begin{enumerate}
                \item \textbf{EM3 - Scalar Bias}: The bias has the same value $b_{x} = b_{y} = b_{z} = b$ in all axes, and thus is addressed as a scalar.
                \item \textbf{EM4 - Vector Bias}: The bias values may differ in each axis as defined in \eqref{bias_vec_def}.
            \end{enumerate}
        \item \textbf{EM5 - Vector Error Model}: Both scale factor and bias are vectors, allowing different combinations of scale factor and bias value for each axis, as presented in \eqref{scale_vec_def} and \eqref{bias_vec_def}.
\end{enumerate}
Figure \ref{error_models_fig} presents the five EMs considered in this work.

\subsection{Proposed DVL Calibration Neural Network}\label{prop_data_method}
\begin{figure*}[!ht]
	\centering
		\includegraphics[width = 0.99\textwidth]{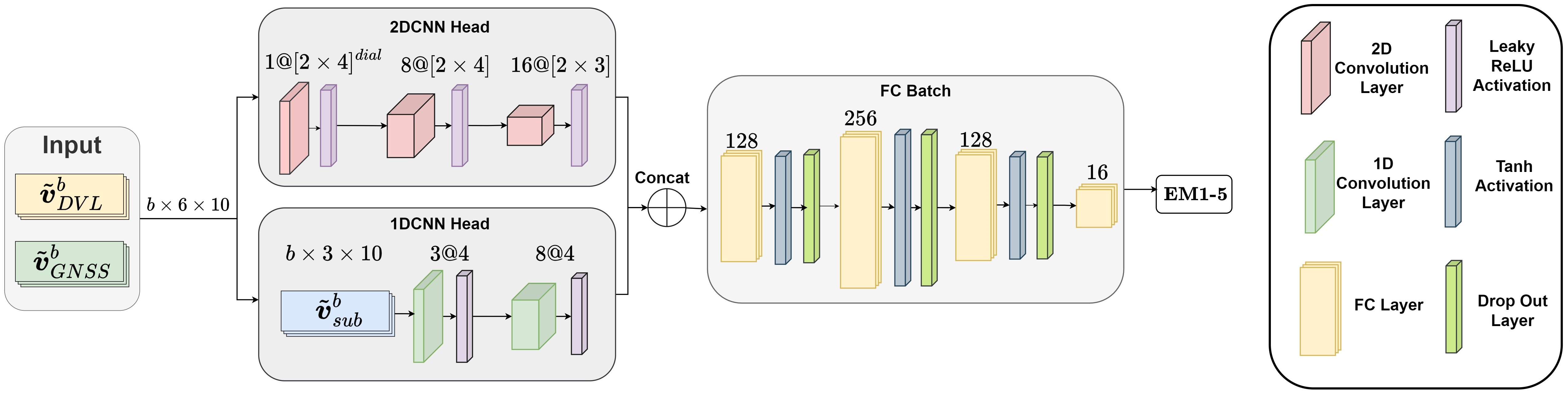}
	  \caption{Block diagram of our DCNet architecture used to estimate one of the proposed \textbf{EM1-EM5} error terms.}\label{prop_nn_arch_diag}
\end{figure*}

\begin{figure}[!ht]
  \begin{subfigure}[c]{0.58\linewidth}
    \centering
    \includegraphics[width=\linewidth]{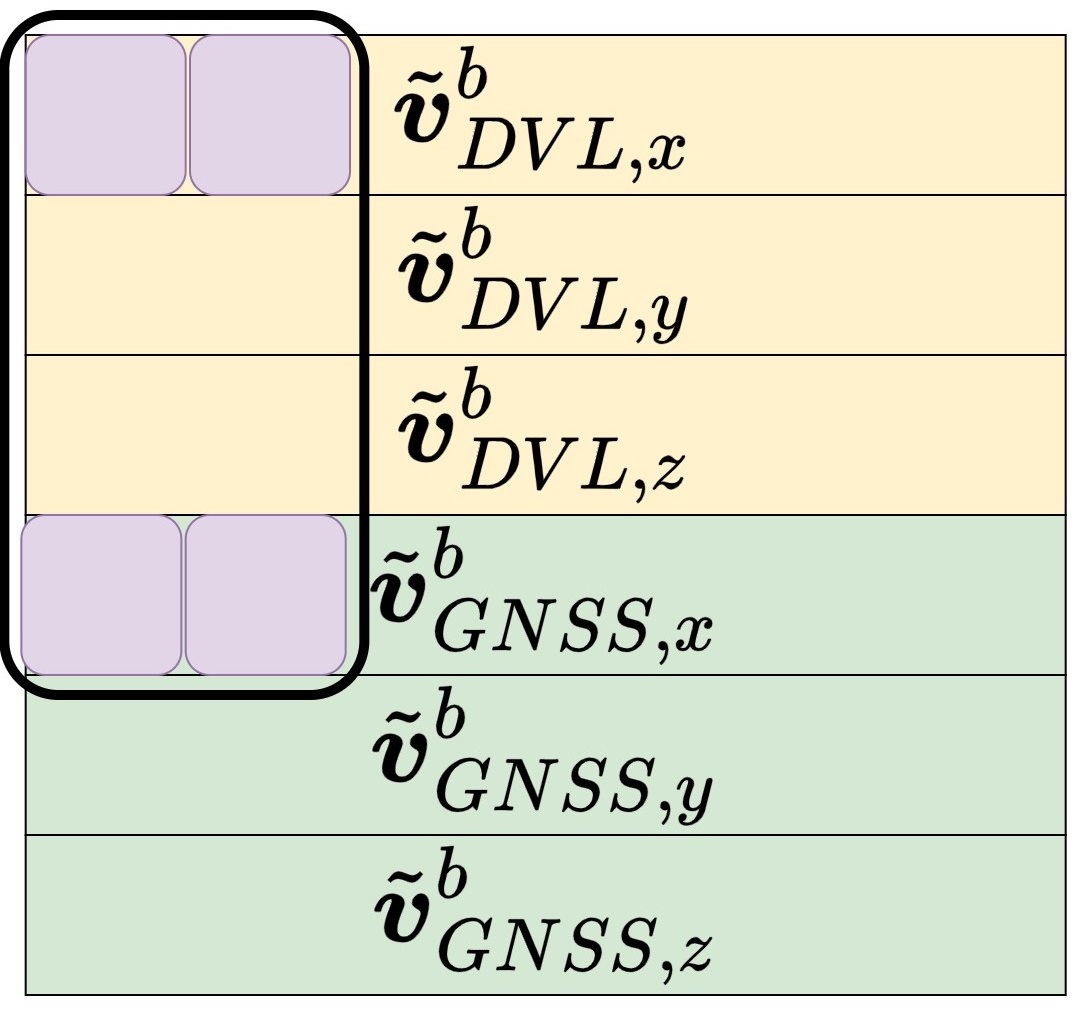}%
    \caption
      {Dilated 2D kernel process both velocity vectors}\label{2d_dil_procc_vels_fig}
  \end{subfigure}\hfill
  \begin{tabular}[c]{@{}c@{}}
    \begin{subfigure}[c]{0.35\linewidth}
      \centering
      \includegraphics[width=\linewidth]{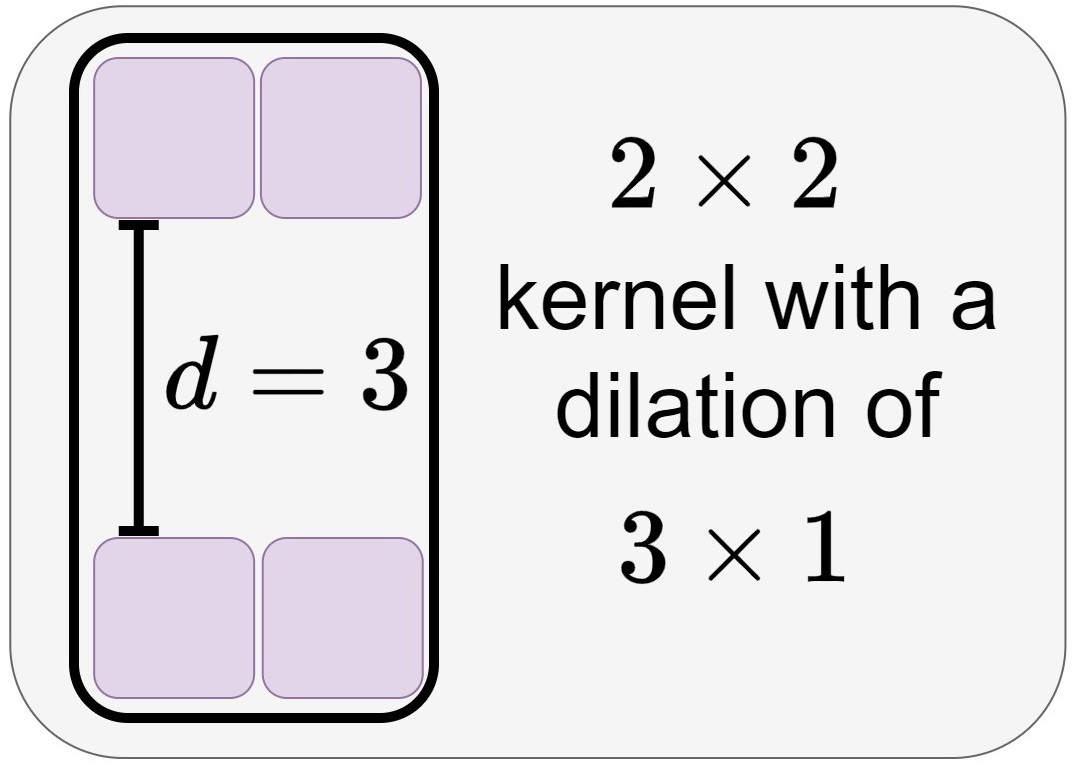}%
      \caption{2D dilated kernel}\label{2d_dil_conv_subfig}
    \end{subfigure}\\
    \noalign{\bigskip}%
    \begin{subfigure}[c]{0.35\linewidth}
      \centering
      \includegraphics[width=\linewidth,page=2]{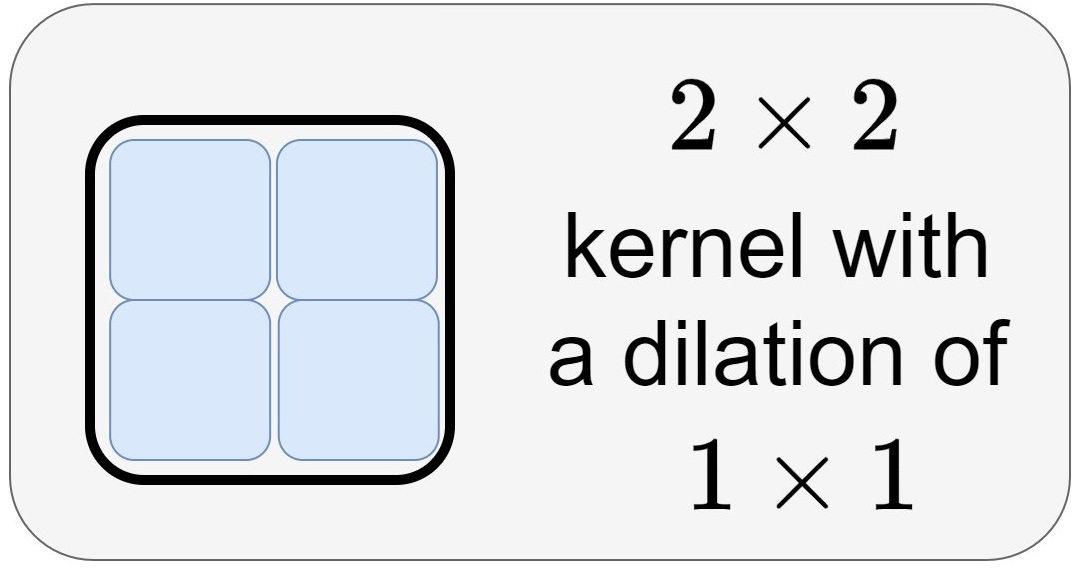}%
      \caption{2D Non-dilated kernel}\label{2d_reg_conv_subfig}
    \end{subfigure}
  \end{tabular}
  \caption{Illustration of two 2D convolution kernels sized $2\times2$ and how the dilated kernel processes both GNSS and DVL velocity vectors. Figure \ref{2d_dil_procc_vels_fig} shows the $3\times1$ dilated 2D convolution kernel processes both X axes of the DVL and GNSS velocities simultaneously. Figure \ref{2d_dil_conv_subfig} illustrates the $3\times1$ dilated 2D kernel, while Figure \ref{2d_reg_conv_subfig} shows a $1\times1$ 2D kernel, which in fact is referred to as not dilated.}\label{dilated_kernel_exmp}%
\end{figure}

\noindent Our proposed approach, DVL calibration neural network (DCNet), is a data-driven approach utilizing a deep learning (DL) neural network (NN) framework to estimate the error terms in the different \textbf{EM1} - \textbf{EM5} as presented in Section \ref{prop_err_model}. Our NN, presented in Figure \ref{prop_nn_arch_diag}, is suitable for all EMs. This is a multi-head network with input, output, and three major parts, as described below:
\begin{enumerate}
    \item \textbf{Input Size}: The input to the network is the same as the input to that presented in \eqref{gnss_dvl_fin_error_model}; that is the DVL and GNSS-RTK velocity vectors that were transformed to body frame, denoted as $\Tilde{\boldsymbol{v}}_{\mathrm{DVL}}^{b}$ and $\Tilde{\boldsymbol{v}}_{\mathrm{GNSS}}^{b}$, respectively. The input size is $<batch\:size \times 6 \times window\:size>$, where 6 is the number of stacked velocity axes, which are the DVL and GNSS-RTK XYZ axes, and 10 is the selected window size that is described in subsection \ref{data_sub_sec}.
    \item \textbf{1DCNN Head}: The input velocity vectors are subtracted to produce $\boldsymbol{v}_{\mathrm{sub}}^{b}$, which has has only three axes, X, Y, and Z. The subtraction vector goes through two 1D convolution layers each followed by a LeakyReLU activation function \eqref{leakyrelu}. The vector $\boldsymbol{v}_{\mathrm{sub}}^{b}$ is the subtraction of the GNSS-RTK from the DVL as follows:
    \begin{equation}\label{v_sub_def_eq}
        \Tilde{\boldsymbol{v}}_{\mathrm{sub}}^{b} =  \Tilde{\boldsymbol{v}}_{\mathrm{DVL}}^{b} -  \Tilde{\boldsymbol{v}}_{\mathrm{GNSS}}^{b}.
    \end{equation}
    We used the subtraction, because, if we consider a scalar scale factor error model such as \textbf{EM1}, we can use \eqref{gnss_dvl_fin_error_model} as follows:
    \begin{equation}\label{initial_sub_vector}
        \Tilde{\boldsymbol{v}}_{\mathrm{sub}}^{b} = (1+\boldsymbol{k}_{\mathrm{DVL}}){\mathbf{R}}_{n}^{b}\boldsymbol{v}^{n} - \Tilde{\boldsymbol{v}}_{\mathrm{GNSS}}^{b} 
    .\end{equation}
    At this point we ignore $\boldsymbol{\sigma v}^{d}$ since it can be averaged out. Recall that we use the GNSS-RTK as the reference and we assume $\hat{\mathbf{R}}_{n}^{b} = \mathbf{I}_{3}$, thus resulting in
    \begin{equation}\label{v_n_is_gnss_assu_eq}
        \hat{\mathbf{R}}_{n}^{b}\boldsymbol{v}^{n} = \Tilde{\boldsymbol{v}}_{\mathrm{GNSS}}^{b}
    .\end{equation}
    By using \eqref{v_n_is_gnss_assu_eq} with \eqref{initial_sub_vector} we derive the following:
    \begin{equation}\label{final_sub_vec}
    \begin{aligned}
        \Tilde{\boldsymbol{v}}_{\mathrm{sub}}^{b} = &(1+\boldsymbol{k}_{\mathrm{DVL}})\Tilde{\boldsymbol{v}}_{\mathrm{GNSS}}^{b}  - \Tilde{\boldsymbol{v}}_{\mathrm{GNSS}}^{b} \\
        & = \Tilde{\boldsymbol{v}}_{\mathrm{GNSS}}^{b} + \boldsymbol{k}_{\mathrm{DVL}} \cdot \Tilde{\boldsymbol{v}}_{\mathrm{GNSS}}^{b} - \Tilde{\boldsymbol{v}}_{\mathrm{GNSS}}^{b} \\
        & = \boldsymbol{k}_{\mathrm{DVL}} \cdot \Tilde{\boldsymbol{v}}_{\mathrm{GNSS}}^{b}
    \end{aligned}
    .\end{equation}
    Considering that we are using several training trajectories, we expect that the FC batch will be able to extract the error terms more easily when provided with the concatenated outputs of the two CNN heads, given the derived input \eqref{initial_sub_vector} to the 1DCNN.
    \item \textbf{2DCNN Head}: Is passed with the two velocity vectors, the GNSS-RTK and the DVLs, as presented in the input, through three 2D convolution layers, each followed by a LeakyReLU activation function \eqref{leakyrelu}. The first convolution's layer kernel is dilated by $3\times1$, while the other two are dilated by $1\times1$. This way, instead of processing the extracted features by the 1D convolution layers only once concatenated and passed through the FC batch, using the 2D convolution kernel allows us to extract features and process information stored in both vectors from the start through the first convolution layer. The dilated kernel in the first 2D convolution layer allows the first layer to process simultaneously the corresponding axes of the DVL and GNSS-RTK velocity; i.e., process both X axes. The motivation for the dilation is to enable the network to mimic the information extraction as in the subtraction vector, $\Tilde{\boldsymbol{v}}_{sub}^{b}$. Figure \ref{dilated_kernel_exmp} illustrates the dilation versus the non-dilated kernel.
    \item \textbf{FC Batch}: consists of four FC layers. Each layer of the first three is followed by a LeakyReLU activation function \eqref{leakyrelu} and then by a dropout layer \ref{regularization_subsec}. The final layer directly outputs the error terms according to the evaluated error model, \textbf{EM1} - \textbf{EM5}. The outputs of both convolution heads are flattened and then stacked, or concatenated over each other to create the input to the FC batch.
    \item \textbf{Output}: The output itself and its size are determined by which error model \textbf{EM1-EM5} is being evaluated; i.e., if \textbf{EM2} is evaluated, the output would be a $\hat{\boldsymbol{s}} \in \mathbb{R}^{3}$.
\end{enumerate}
Next, we provide a mathematical formulation of our DCNet, describe the training process, and provide the implementation details.
All the NN hyper parameters such as the batch size, the dropout probability, and activation function formulations are described in Section \ref{training_process_section} and the values are presented in Table \ref{hyper_params_table}.
\subsection{DCNet Formulation and Training Process}\label{training_process_section}
\noindent In this section, we will examine the mathematical formulation of DCNet and the training process.
\subsubsection{Fully Connected Neural Network}\label{FC_subsec}
\noindent Neurons are the building blocks of a NN, where each neuron in the FC layer is defined by \cite{cohen2022beamsnet}:
\begin{equation}\label{neuron_def}
    {z}_{i}^{(l)} = \sum_{j=1}^{n_{(l-1)}} {{\omega}_{ij}^{(l)}{a}_{j}^{(l-1)}} + {b}_{i}^{(l)}
\end{equation}
where ${\omega}_{ij}^{(l)}$ is the learnable weight of the $i^{th}$ neuron in the $l^{th}$ layer that is connected to the $j^{th}$ neuron in the $(l-1)^{th}$ layer. The term ${b}_{i}^{(l)}$ is the bias of the $i^{th}$ neuron in the $l^{th}$ layer, and ${a}_{j}^{(l-1)}$ is the output of the $j^{th}$ neuron in the $(l-1)^{th}$ layer. Each convolution layer is structured in a similar way where each kernel sized ${m}_{1} \times {m}_{2}$ contains ${m}_{1} \times {m}_{2}$ neurons.
\subsubsection{Convolution Neural Network}\label{cnn_subsec}
\noindent CNNs are constructed similarly to a FC layer, and expand the single neuron \eqref{neuron_def} to a kernel (filter) of size ${m}_{1}\times {m}_{2}$ neurons \cite{cohen2022beamsnet}:
\begin{equation}\label{conv_kernel_eq_def}
    C_{ij}^{(l)} = \sum_{\alpha=0}^{m_1} \sum_{\beta=0}^{m_2} {{\omega}_{\alpha\beta}^{(r)}{a}_{(i+\alpha)(i + \beta)}^{(l-1)}} + {b}^{(r)}
\end{equation}
where $C_{ij}^{(l)}$ is the output of the kernel, and ${\omega}_{\alpha\beta}$ is the weight of the kernel in the $(\alpha , \beta)$ position in the $r^{th}$ convolution layer. The term ${a}_{(i+\alpha)(i + \beta)}^{(l-1)}$ is the output of the preceding convolution layer, and ${b}^{(r)}$ is the bias of the $r^{th}$ convolution layer.
\subsubsection{Nonlinear Activation Functions}\label{activation_func_subsec}
\noindent Traditionally, each neuron's output, whether of a FC layer \eqref{neuron_def} or a convolution layer \eqref{conv_kernel_eq_def}, passes through a nonlinear activation function, otherwise a NN acts as a linear regression model \cite{sharma2017activation}:
\begin{equation}\label{general_act_func_eq}
    a_{i}^{(l)} = h({z}_{i}^{(l)})
\end{equation}
where ${z}_{i}^{(l)}$ is the $i^{th}$ neuron in the $l^{th}$ layer output---in this case a FC layer \eqref{neuron_def}---$h$ is the activation function, and $a_{i}^{l}$ is the output of the activation function employed. In this work we use the following nonlinear activation functions:
\begin{enumerate}
    \item \textbf{Leaky Rectified Linear Unit (LeakyReLU)}: An activation function that scales negative values and does not change positive values \cite{sharma2017activation}:
    \begin{equation}\label{leakyrelu}
        LeakyReLU(C_{ij}^{(l)}x) = 
        \begin{cases}
                C_{ij}^{(l)},& \text{if } C_{ij}^{(l)} \geq 0\\
                \alpha \cdot C_{ij}^{(l)},              & \text{otherwise}
                \end{cases}
    \end{equation}
    where $\alpha$ is the scaling factor of the negative value, and $C_{ij}^{(l)}$ is the input to the function, in this case a convolution layer kernel output \eqref{conv_kernel_eq_def} since LeakyReLU is applied over the convolution layers output both in the 1DCNN and the 2DCNN heads.
    \item \textbf{Hyperbolic Tangent Function (Tanh)}: A continuous and differentiable nonlinear activation function, which outputs values within a defined range of $[-1,1]\in\mathbb{R}$ as in \cite{sharma2017activation}:
    \begin{equation}\label{tanh_eq}
        Tanh({z}_{i}^{(l)}) = \frac{e^{{z}_{i}^{(l)}} - e^{-{z}_{i}^{(l)}}}{e^{{z}_{i}^{(l)}} + e^{-{z}_{i}^{(l)}}}
    \end{equation}
    where $e$ is the Euler constant, and ${z}_{i}^{(l)}$ is the FC layer output, which is the input to this activation function. As described in Section \ref{prop_approach}, the TanH activation function is applied over the first three layers in the FC batch.
\end{enumerate}
\subsubsection{Regularization}\label{regularization_subsec}
\noindent During training, overfitting occurs when the NN learns noise patterns present in training data, which results in a low loss value during training but fails to generalize and performs poorly on the test data \cite{salman2019overfitting}. In this work, we use dropout regularization \cite{srivastava2014dropout} to mitigate the overfitting problem. With dropout, some neurons are not used during training with probability $1-p$, as follows \cite{srivastava2014dropout}:
\begin{equation}
\begin{split}
    r_{i}^{l} & \sim Bermoulli(p) \\
    \Tilde{a}_{i}^{l} & = r_{i}^{l} \cdot a_{i}^{l} \\
    a_{j}^{l+1} & = {{\omega}_{ji}^{(l+1)}\Tilde{a}_{i}^{(l)}} + {b}_{j}^{(l+1)}
\end{split}
\end{equation}
where $r_{i}^{l}$ is a random variable sampled from a $Bernoulli$ distribution with $p$ probability, $a_{i}^{l}$ is the activation function output over the $i^{th}$ neuron \eqref{general_act_func_eq}, and the term $\Tilde{a}_{i}^{l}$ is the dot product of $r_{i}^{l}$ and $a_{i}^{l}$. The dot product, $\Tilde{a}_{i}^{l}$, is fed as input to the $j^{th}$ neuron in the $(l+1)^{th}$ layer after dropout was applied. The terms ${\omega}_{ji}^{(l+1)}$ and ${b}_{j}^{(l+1)}$ are the weights and biases as described in Section \ref{FC_subsec}.
\subsubsection{Training Process}\label{training_process_loss_subsec}
\begin{figure}[!ht]
	\centering
		\includegraphics[width = 0.95\columnwidth]{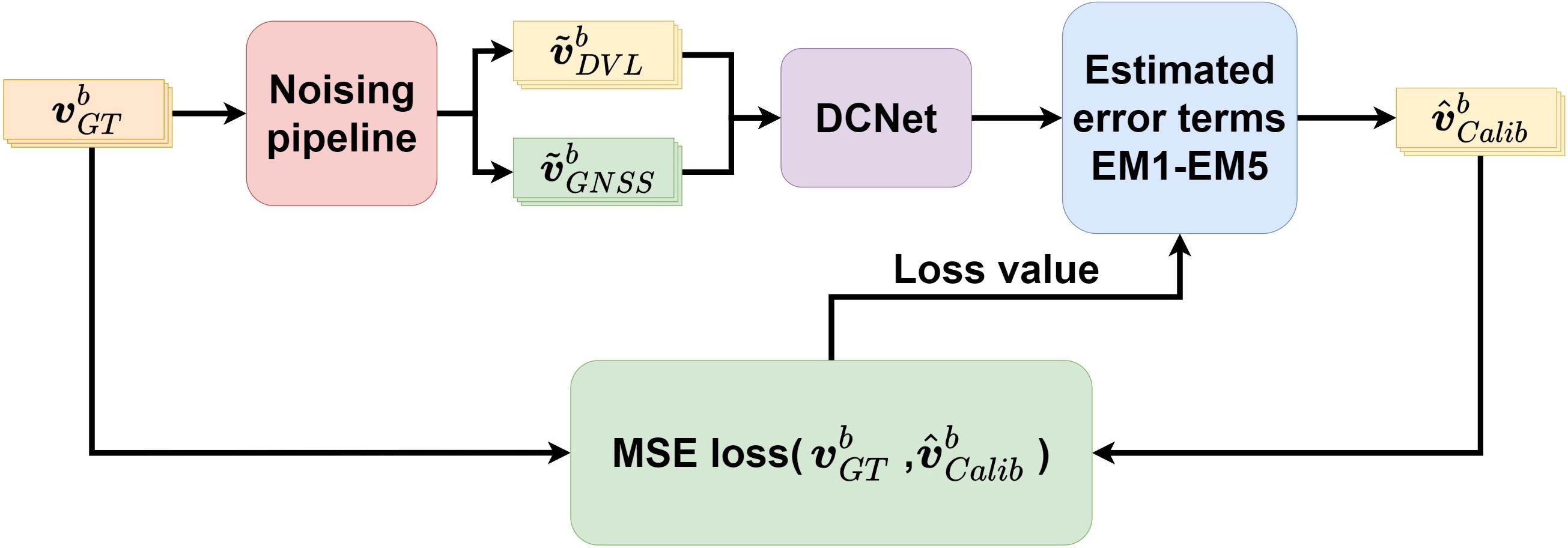}
	  \caption{Block diagram showing the closed loop MSE loss calculation procedure after the proposed approach estimates the error terms.}\label{mse_loss_fig}
\end{figure}

\noindent The training process aims to determine the weights and biases of all the neurons comprising the NN, in order to achieve the lowest possible loss function score. The loss function used in this study is the mean squared error (MSE) loss function:
\begin{equation}\label{mse_eq}
    J(\boldsymbol{y}_i, \hat{\boldsymbol{y}}_i) = \frac{1}{N} \mid\mid \boldsymbol{y}_i - \hat{\boldsymbol{y}}_i \mid\mid ^2
\end{equation}
where $J()$ is the MSE function, $\boldsymbol{y}_i$ is the GT value, $\hat{\boldsymbol{y}}_i$ is the network estimation, and $N$ is the number of samples. Forward passing refers to the process of passing the input through the network's various layers, such as \eqref{neuron_def} and \eqref{conv_kernel_eq_def}. Based on the output of the network, the loss function value is calculated and derived with respect to the weights and biases of the NN. As part of the training process, the derivations are used to minimize the loss value in the next forward pass, by updating the weights and biases of the model based on those derivations. To accomplish this, a gradient descent algorithm is used:  
\begin{equation}\label{grad_desc_eq}
    \boldsymbol{\theta} = \boldsymbol{\theta} - \eta\nabla_{\theta}J(\theta), \: \:\:\: \boldsymbol{\theta} = [\omega \: \: \: b]^T
\end{equation}
where $\boldsymbol{\theta}$ is the vector of the weights and biases of a neuron, $\eta$ is the learning rate that is set at the start of the training process, $J(\boldsymbol{\theta})$ is the loss function value with respect to the vector of the weights and biases $\boldsymbol{\theta}$, and $\nabla_{\theta}$ is the gradient operator with respect to the vector $\boldsymbol{\theta}$.
To this end, a root mean squared propagation (RMSProp) optimizer is used \cite{mukkamala2017variants}. We set the learning rate of the proposed NN to $5\cdot 10^{-5}$ when training it to estimate the error terms of \textbf{EM1-EM4}, and to $5 \cdot 10^{-4}$ when training it to estimate the error terms of \textbf{EM5}. Note that the network estimates the error terms for one of \textbf{EM1-EM5} based on the error model that is being evaluated. The loss value, however, is calculated on the basis of velocities rather than error terms themselves. Following the network estimation of the error terms, the input DVL velocity is calibrated using the estimated error terms, and using both the input and calibrated values, the calculated loss value is used to update the weights. Accordingly, the MSE loss using two velocities is as follows:
\begin{equation}\label{mse_loss_vel_eq}
    \centering
    MSE(\boldsymbol{y}_{i} , \hat{\boldsymbol{y}_{i}}) = \frac{\sum_{i=1}^{N} [\sum_{j}^{\{X,Y,Z\}}(\boldsymbol{y}_{i,j} - \hat{\boldsymbol{y}}_{i,j})^{2}]} {N}
\end{equation}
where $\boldsymbol{y}_{i} \in \mathbb{R}^{N\times3}$ is the GT velocity vector, $\hat{\boldsymbol{y}}_{i} \in \mathbb{R}^{N\times3}$ is the calibrated velocity vector, $i$ indicates the length of the velocity vector---that is, $N$---and $j$ is the three axes X, Y, and Z. Figure \ref{mse_loss_fig} illustrates a block diagram of the forward pass and the input to the closed loop MSE loss function.
\\
For the purpose of this section---that is, the weights and biases updated during training---the output of the MSE loss function is sufficient. 
To prevent the proposed NN approach from overfitting the data, a dropout regularization is used between the first three layers in the FC head as presented in Figure \ref{prop_nn_arch_diag}.
All the hyperparameters for the training procedure described in Section \ref{prop_data_method} and fully formulated in this section are presented in Table \ref{hyper_params_table}.
\begin{table}[!ht]
\caption{Hyperparameters used during the training process of our proposed NN approach.}\label{hyper_params_table}
\centering
\begin{adjustbox}{width = \columnwidth}
\begin{tabular}{|c|c|c|c|c|c|}
\hline
Parameter & Epochs & B. size & 2D Leaky & 1D Leaky & Dropout p  \\ \hline
Value     & 100    & 256     & 0.05     & 0.05     & 0.3        \\ \hline
\end{tabular}
\end{adjustbox}
\end{table}
\section{Experimental Results}\label{res_sec}
\subsection{Data Collection and Processing}\label{data_sub_sec}
\begin{figure}[!ht]
	\centering
		\includegraphics[width = 0.8\columnwidth]{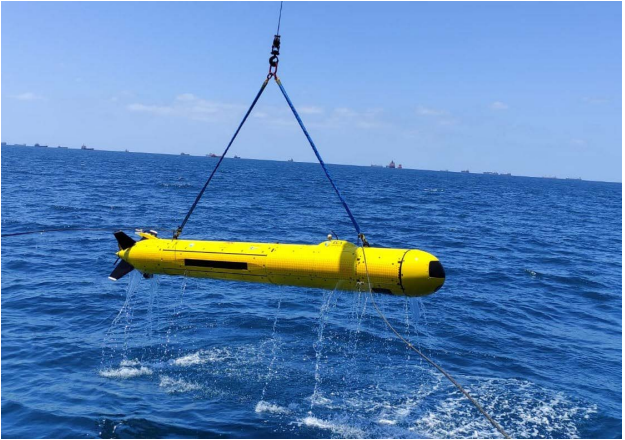}
	  \caption{The "Snapir" AUV being pulled out of the water after a mission.}\label{snapir_auv_image}
\end{figure}

\begin{figure}[!ht]
    \centering
    \begin{subfigure}[t]{0.85\columnwidth}
        \includegraphics[width=1\linewidth]{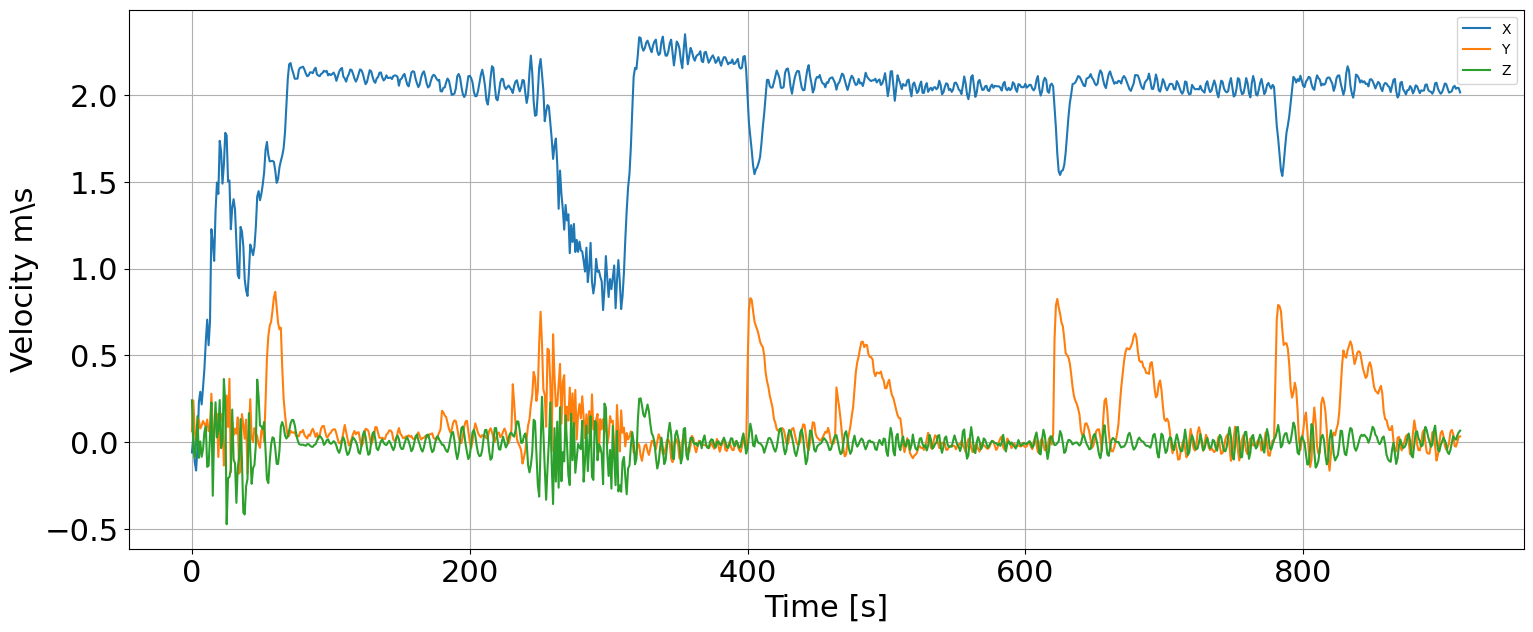}
        \caption{T4 Velocity Recording}
        \label{mission_4_v_in_body_fig} 
    \end{subfigure}
    \begin{subfigure}[b]{0.85\columnwidth}
        \includegraphics[width=1\linewidth]{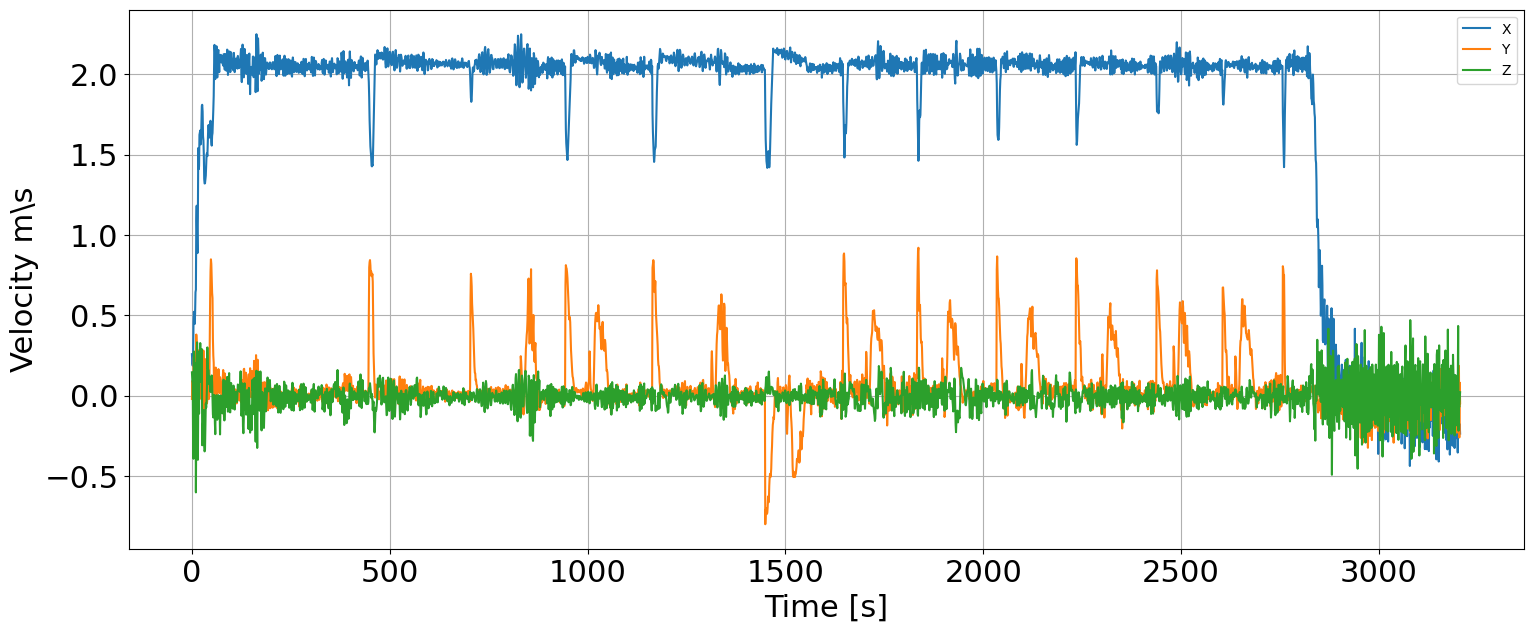}
        \caption{T6 Velocity Recording}
        \label{mission_6_v_in_body_fig}
    \end{subfigure}
\caption{AUV velocity as measured by the DVL during T4 and T6 trajectories.}\label{missions_4_6_v_body_figs}
\end{figure}

\begin{figure}[!ht]
	\centering
		\includegraphics[width = 0.95\columnwidth]{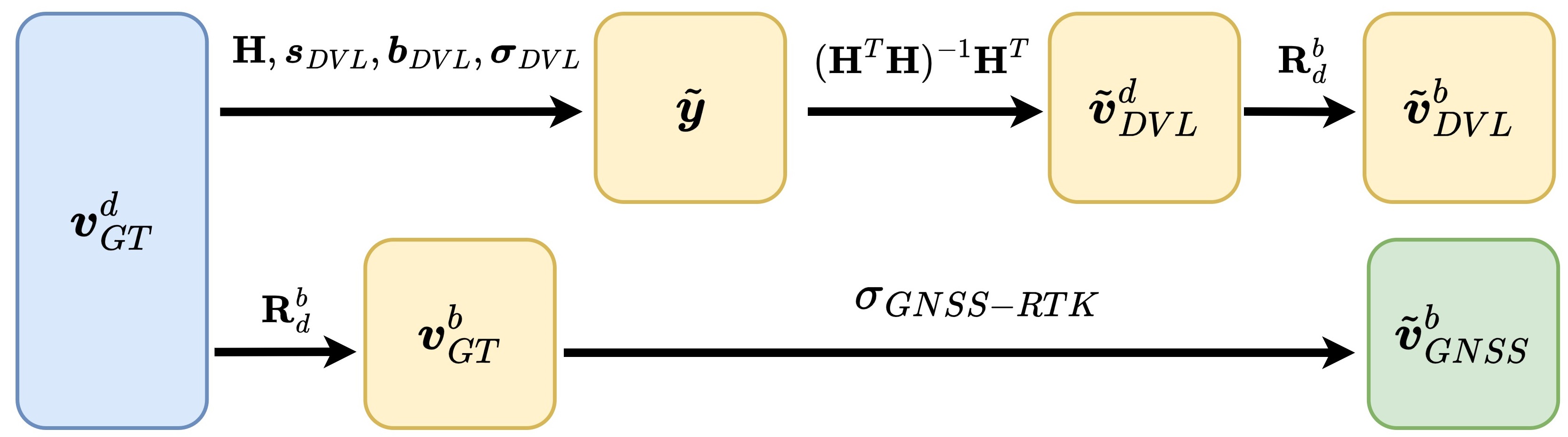}
	  \caption{The pipeline used to generate   noisy GNSS and DVL velocity measurements based on the GT velocity from the sea experiments.}\label{data_pipeline_fig}
\end{figure}

\noindent To validate our proposed approach, data recorded using the University of Haifa's "Snapir" AUV was used. This data was provided in \cite{cohen2022beamsnet} and \cite{shurin2022autonomous} and can be found at their associated GitHub repositories. Figure \ref{snapir_auv_image} shows an image of the AUV during a mission. "Snapir" is a torpedo-shaped AUV measuring 5.5 meters in length and 0.5 meters in diameter. The "Snapir" employs a Teledyne RD Instruments Work Horse Navigator DVL, with a sampling rate of 1Hz and accuracy of 0.6 cm/s while traveling at 3 m/s \cite{levy2023ins}.
\\
A total of ten trajectories were recorded in two separate sea experiments. In the first, nine trajectories, referred to as T1 - T9, were recorded with a total time of 243 minutes. In the second experiment, an additional trajectory was recorded (referred to as T10) with a total time of 33 minutes of recordings. Figure \ref{missions_4_6_v_body_figs} presents the AUV velocity, expressed in the body frame, as a function of time during the T4 and T6.\\
Next, trajectory T10 was divided into two parts. The first 200 seconds were allocated for evaluating the calibration parameters based on our trained network and are referred to as the calibration dataset T10Cal. The remaining 1800 seconds of T10 are part of the test dataset and are referred to as T10Test. The training dataset contains of trajectories T1-T5 while the test datasets includes trajectories T6-T9 and T10Test. The motivation for this selection is twofold: 1) T10 is the trajectory recorded in a different sea experiment, thus making it the calibration dataset aims to generalize and increase our model robustness as we train on trajectories from a different sea experiment. 2) T10Cal represents a simple nearly constant velocity trajectory, thus if we are able to learn the calibration parameters solely on T10Cal, we also offer a simple effective trajectory for the calibration process. 
\\
\noindent All the recordings contain the DVL measured velocity. To produce the sensor noise, we applied a moving average (MA) filter over all ten trajectories. We used a MA filter with a five second window \cite{gonzalez2018statistical}:
\begin{equation}\label{ma_filter_form}
    \boldsymbol{x}_{i}^\prime = \frac{1}{T} \sum_{j=i}^{i+5} \boldsymbol{x}_{j}
\end{equation}
where $T$ is the total recording length, $\boldsymbol{x}_{j}$ is the original velocity vector at time step $j$, and $\boldsymbol{x}_{i}^\prime$ is the MA filtered velocity vector. The filtered data is then referred to as our ground-truth (GT) data.\\
To produce the low-cost DVL measurements and the reference GNSS-RTK, we employed a noising pipeline presented in Figure \ref{data_pipeline_fig}, which takes as input the GT velocity. The noising pipeline is designed to mimic the beams error model \eqref{y_as_func_of_H_and_v_error_model}, transformation to the DVL frame \eqref{psudo_inverse}, and later to the body frame. Additionally, the pipeline outputs the GNSS-RTK measurements by applying a zero mean Gaussian white noise to the GT measurements as follows:
\begin{equation}\label{gnss_rtk_noising_eq}
\Tilde{\boldsymbol{v}}_{\mathrm{GNSS}}^{b} = \boldsymbol{v}_{\mathrm{GT}}^{b} + \boldsymbol{\sigma}_{\mathrm{GNSS-RTK}}
\end{equation}
where $\boldsymbol{\sigma}_{\mathrm{GNSS-RTK}} \sim N(0,{0.005\mathrm{m/s}}^2)$ and remains constant whenever the pipeline is used. Note that $s_{\mathrm{DVL}}$, ${b}_{\mathrm{DVL}}$, and ${\sigma}_{\mathrm{DVL}}$ are the only parameters that affect the noising pipeline outcome.
\\
To create a sufficient sized train dataset that captures a wide range of DVL measurements characteristics, 1134 combinations of the error terms were used with the pipeline. Table \ref{train_traj_params_tbl} presents the error terms values and combinations that were added to the train trajectories T1 - T5. After adding the error terms, the trajectories were split into windows of 10 seconds with a stride of 9 seconds, resulting in 835,758 data points, each of size $6 \times 10$, where 6 is the stacked XYZ axes of the noised DVL and GNSS-RTK velocities. Then, the data points were split by a 80:20 $[\%]$ ratio, which resulted in 669,060 data points for the train set and 166,698 data points for the evaluation set.\\
To evaluate our proposed calibration approach, two error terms combinations were selected to noise the five test trajectories, T6 - T10. Table \ref{dvl_types_under_test} presents the two error terms combination selected, which are referred to as DVL1 and DVL2. Both DVL1 and DVL2 error terms were selected to mimic lower-grade DVLs.
\begin{table}[!ht]
    \caption{DVL beam error model parameters used to create the training set.}\label{train_traj_params_tbl}
    \centering
    \begin{adjustbox}{width=\columnwidth,center}
    \begin{tabular}{c|c|c|c|c|}
        \cline{2-5}
                                 & Lower Value & Upper Value & Step size & $\#$ Values \\ \hline
        \multicolumn{1}{|c|}{$\boldsymbol{s}_{\mathrm{DVL}}$ $[\%]$}      & 0.2        & 1.5         & 0.1       & 14           \\ \hline
        \multicolumn{1}{|c|}{$\boldsymbol{b}_{\mathrm{DVL}}$ [cm/s]}       & 0.1      &    0.9    & 0.1     & 9            \\ \hline
        \multicolumn{1}{|c|}{$\boldsymbol{\sigma}_{\mathrm{DVL}}$ [cm/s]}  & 0.01     & 0.1       & 0.01    & 9            \\ \hline
    \end{tabular}
    \end{adjustbox}
\end{table}
\begin{table}[!ht]
    \caption{DVL beam error model parameters used to create the testing set. Noise values are not present in the train parameters.}\label{dvl_types_under_test}
    \centering
    \begin{adjustbox}{width=\columnwidth,scale =0.9}
    \begin{centering}
    \begin{tabular}{c|c|c|c|}
    \cline{2-4}
           & Scale $[\%]$ & Bias [cm/s] & DVL Noise [cm/s]\\ \hline
        \multicolumn{1}{|c|}{DVL 1} & 1.0   & 0.7 & 2.0      \\ \hline
        \multicolumn{1}{|c|}{DVL 2}  & 1.0   & 0.7 & 0.02    \\ \hline
    \end{tabular}
    \end{centering}
    \end{adjustbox}
\end{table}
\subsection{Evaluation Procedure}\label{eval_procedure}
\noindent This section describes the evaluation procedure, which consists of calibration and evaluation phases. The procedure is designed to resemble a real-life calibration process, whereby the AUV follows a predefined calibration trajectory for a predetermined period of time, after which the error terms are estimated. Thus, the following stages were designed.\\
We employed the 200-second T10Cal dataset for the calibration phase. It serves a double purpose: first, the calibration error terms are estimated; secondly, the convergence time of each approach is also determined. We examined five different calibration window sizes to find the one that produces the best performance. The windows sizes are $w_i = 20, 40, 60, 80, 100$ seconds. The calibration takes place only during this window. Once the DVL calibration parameters are estimated, they are applied on the velocity measurements during the rest of the calibration phase. For example, for a 40-second window size, calibration parameters are estimated. The velocity is calibrated for the rest of the 160 seconds using the parameters estimated for the calibration during the first 40 seconds.\\
The accuracy of the error terms in each window for each approach is estimated by calculating the root mean squared error (RMSE) of the calibrated velocity of the remaining time against the GT, as follows:
\begin{equation}\label{rmse_vel_eq}
    \centering
    \mathrm{RMSE}(\boldsymbol{x}_{i} , \hat{\boldsymbol{x}_{i}}) = \sqrt{\frac{\sum_{i=1}^{N} \sum_{j = X,Y,Z}(\boldsymbol{x}_{i,j} - \hat{\boldsymbol{x}}_{i,j})^{2}} {N}}
\end{equation}
where $\boldsymbol{x}_{i}$ is the GT velocity vector, $\hat{\boldsymbol{x}_{i}}$ is the calibrated velocity vector, and $N$ is the length of the trajectory in seconds. Once all calibration window sizes have been examined, the selected calibration parameters for each error model and the baseline approach are those that achieved the lowest RMSE \eqref{rmse_vel_eq}. The corresponding window size is referred to as the calibration convergence time.\\
Once the calibration phase concludes, we evaluate the performance on the test dataset using the estimated calibration parameters and \eqref{rmse_vel_eq}. That is, we use DCNet estimated parameters for all five error models \textbf{EM1}-\textbf{EM5}. For the baseline approach, which calculates only the scale factor, we use \eqref{average_direct_scale}. We examine DV1 and DVL2 parameters for the baseline and for \textbf{EM1}-\textbf{EM5}.
\subsection{Results}\label{results_sub_sec}
\begin{figure}[!ht]
	\centering
		\includegraphics[width = 0.7\columnwidth]{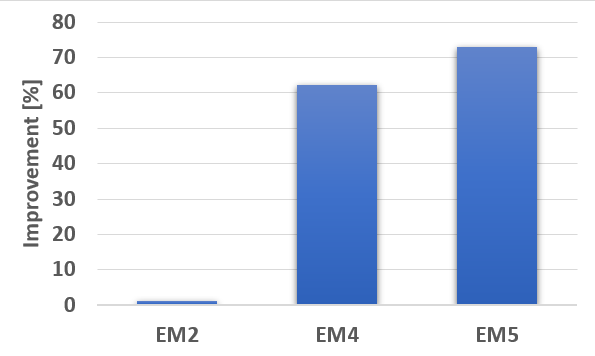}
	  \caption{RMSE at $\mathrm{T_{conv}} = 20$ seconds of DCNet employing \textbf{EM2}, \textbf{EM4}, and \textbf{EM5}, evaluated on DVL2 for dataset T10Cal.}\label{calib_improv_in_prec}
\end{figure}

\begin{figure}[!ht]
	\centering
		\includegraphics[width = 0.75\columnwidth]{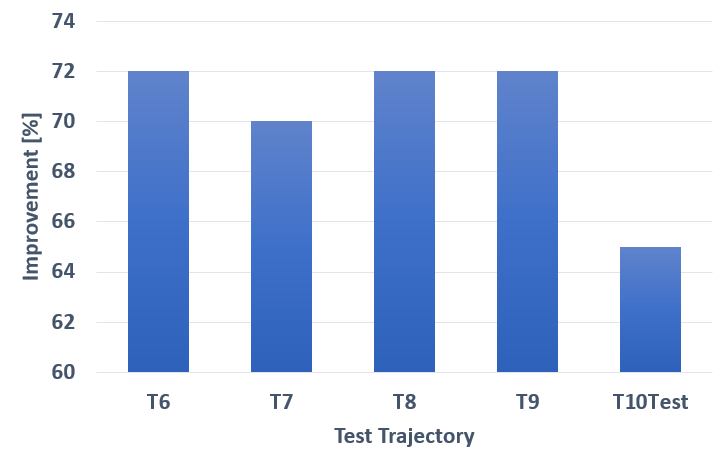}
	  \caption{Improvement [\%] in RMSE of DCNet employing \textbf{EM5} over the baseline approach with DVL2.}\label{prop_nn_em5_improv_BL_dvl2_eval_traj}
\end{figure}

\noindent To verify the accuracy of the evaluation procedure on each EM and minimize the effect of outliers, the evaluation procedure is repeated for 200 Monte Carlo (MC) iterations. For each run, the calibration parameters were evaluated in the different window sizes for each EM, and then the RMSE \eqref{rmse_vel_eq} was calculated for the calibration dataset T10Cal. The convergence time is the window size based upon the minimum RMSE that was achieved. That is:
\begin{equation}\label{t_conv_def}
    \mathrm{T_{conv}} = min(\mathrm{RMSE}_{20},\ldots,\mathrm{RMSE}_{100})
\end{equation}
where the subscript of $\mathrm{RMSE}$ is the calibration window time in seconds. Note that the convergence time \eqref{t_conv_def} is the required time for the calibration process. The results of the calibration phase are given in Table \ref{calib_rmse_res_tbl} showing $\mathrm{T_{conv}}$ and its associated RMSE.
\begin{table}[!ht]
\caption{RMSE [cm/s] and $\mathrm{T_{conv}}$ [second] for \textbf{EM1} - \textbf{EM5} using DCNet and the baseline approach during the calibration phase (T10Cal).}\label{calib_rmse_res_tbl}
\centering
\begin{adjustbox}{width = 0.99\columnwidth}
\begin{tabular}{|c|c|c|c|c|c|c|}
\hline
    {}& Baseline & \textbf{EM1} & \textbf{EM2}  & \textbf{EM3}  & \textbf{EM4}   & \textbf{EM5}             \\ \hline
DVL1 & 5.94,(100)  & 5.95,(100) & 5.94,(100) & 6.29,(100) & 5.96,(100) & \textbf{5.90,(40)}        \\ \hline
DVL2 & 0.74,(100)  & 0.74,(100) & \textbf{0.73,(20)} & 2.07,(100) & \textbf{0.28,(20)} & \textbf{0.2,(20)} \\ \hline
\end{tabular}
\end{adjustbox}
\end{table}\\
When considering the noising characteristics of DVL1, DCNet outperformed the baseline approach only in \textbf{EM5}, both in accuracy and in convergence time. In the latter, 60\% improvement was achieved. When considering DVL2, DCNet improved on the baseline approach in accuracy and convergence time employing \textbf{EM2}, \textbf{EM4}, and \textbf{EM5}. All three EMs achieved $80\%$ improvement in the convergence time, while \textbf{EM5} achieved the best accuracy, improving the RMSE by $72\%$ over the baseline. Figure \ref{calib_improv_in_prec} shows the RMSE improvement at $\mathrm{T_{conv}} = 20$ seconds for DCNet with \textbf{EM2}, \textbf{EM4}, and \textbf{EM5} evaluated on DVL2.
\\
Next, we evaluate the performance of our approach and the baseline on the test dataset. To this end, we calibrate the five test trajectories---T6 - T9 and T10Test---using the estimated error terms by DCNet employing \textbf{EM1} - \textbf{EM5}, and the baseline approach, for both DVL types. The RMSE value of the calibrated test trajectories was calculated against their GT. Table \ref{eval_phase_nn_baseline_acc_and_improv_res} shows the RMSE value of all evaluated approaches, and shows the improvement of DCNet over the baseline approach.
\begin{table*}[!t]
\caption{RMSE results of our proposed NN during the evaluation phase, that is over all the five test trajectories that were noised using the two error terms combinations. The bold font represents the lowest RMSE in each test trajectory.}\label{eval_phase_nn_baseline_acc_and_improv_res}
\centering
\begin{adjustbox}{width = 0.99\textwidth}
\begin{tabular}{|cccccccc|}
\hline
\multicolumn{8}{|c|}{RMSE [cm/s] and Improvement $[\%]$ over the baseline}  \\ \thickhline
\multicolumn{1}{|c|}{\multirow{2}{*}{\backslashbox{DVL type}{Approach}}} & \multicolumn{1}{c|}{\multirow{2}{*}{\begin{tabular}[c]{@{}c@{}}Eval.\\ Traj.\end{tabular}}} & \multicolumn{1}{c|}{\multirow{2}{*}{Baseline}} & \multicolumn{5}{c|}{DCNet}                                                              \\ \cline{4-8} 
\multicolumn{1}{|c|}{}                              & \multicolumn{1}{c|}{}                                                                       & \multicolumn{1}{c|}{}                          & \textbf{EM1} & \textbf{EM2} & \textbf{EM3} & \textbf{EM4} & \textbf{EM5}                 \\ \thickhline
\multicolumn{1}{|c|}{\multirow{5}{*}{DVL1}}         & \multicolumn{1}{c|}{T6}                                                            & \multicolumn{1}{c|}{5.96}                      & 5.96                          & 5.96                          & 6.27                           & 5.99                          & \textbf{5.91 (\textless{}1)} \\ \cline{2-8} 
\multicolumn{1}{|c|}{}                              & \multicolumn{1}{c|}{T7}                                                            & \multicolumn{1}{c|}{5.95}                      & 5.95                           & 5.95                          & 6.22                          & 5.99                          & \textbf{5.91 (\textless{}1)} \\ \cline{2-8} 
\multicolumn{1}{|c|}{}                              & \multicolumn{1}{c|}{T8}                                                            & \multicolumn{1}{c|}{5.93}                      & 5.93                           & 5.93                           & 6.25                           & 5.96                          & \textbf{5.89 (\textless{}1)} \\ \cline{2-8} 
\multicolumn{1}{|c|}{}                              & \multicolumn{1}{c|}{T9}                                                            & \multicolumn{1}{c|}{5.95}                      & 5.95                         & 5.95                           & 6.26                           & 5.99                           & \textbf{5.91 (\textless{}1)} \\ \cline{2-8} 
\multicolumn{1}{|c|}{}                              & \multicolumn{1}{c|}{T10}                                                            & \multicolumn{1}{c|}{5.94}                      & 5.94                           & 5.94                           & 6.18                           & 6.07                          & \textbf{5.90 (\textless{}1)} \\ \thickhline
\multicolumn{1}{|c|}{\multirow{5}{*}{DVL2}}         & \multicolumn{1}{c|}{T6}                                                            & \multicolumn{1}{c|}{0.74}                      & 0.74                           & 0.73 (1)                          & 1.96                          & 0.51 (31)                         & \textbf{0.21 (72)}           \\ \cline{2-8} 
\multicolumn{1}{|c|}{}                              & \multicolumn{1}{c|}{T7}                                                            & \multicolumn{1}{c|}{0.74}                      & 0.74                           & 0.74                           & 1.78                           & 0.57 (23)                         & \textbf{0.22 (70)}           \\ \cline{2-8} 
\multicolumn{1}{|c|}{}                              & \multicolumn{1}{c|}{T8}                                                            & \multicolumn{1}{c|}{0.74}                      & 0.74                          & 0.73 (1)                          & 1.96                          & 0.55 (26)                         & \textbf{0.21 (72)}           \\ \cline{2-8} 
\multicolumn{1}{|c|}{}                              & \multicolumn{1}{c|}{T9}                                                            & \multicolumn{1}{c|}{0.74}                      & 0.74                           & 0.73 (1)                          & 1.91                           & 0.63 (15)                         & \textbf{0.21 (72)}           \\ \cline{2-8} 
\multicolumn{1}{|c|}{}                              & \multicolumn{1}{c|}{T10}                                                            & \multicolumn{1}{c|}{0.74}                      & 0.74                           & 0.74                         & 1.70                          & 1.26                          & \textbf{0.26 (65)}           \\ \hline
\end{tabular}
\end{adjustbox}
\end{table*}\\
For DVL1, DCNet achieved only slightly better RMSE than the baseline approach using \textbf{EM5}, which is less than a 1\% improvement. However, when considering DVL2, DCNet improved on the baseline approach by estimating error terms of \textbf{EM4} and \textbf{EM5}. Using \textbf{EM4}, DCNet improved only on test trajectories T6 - T9, which are from the same sea experiment as the training set, but not on T10Test, which is from a different sea experiment. For \textbf{EM5}, DCNet significantly improved the RMSE over the baseline approach and over all five test trajectories by an average of 70\%. Figure \ref{prop_nn_em5_improv_BL_dvl2_eval_traj} shows DCNet RMSE improvement [\%] over the baseline while employing \textbf{EM5} on DVL2.
\section{Conclusion}\label{conc_sec}
\noindent This work proposes and examines the benefits of data-driven approaches in the DVL calibration process. Most works in the literature require complex trajectories accompanied by model-based nonlinear estimation filters for such a calibration task. In this work we offer a data-driven framework with a simple, yet efficient, calibration network called DCNet. We suggested five different calibration error models for the evaluation over two different types of DVL. For training the network we used a dataset of 175 minutes while the test dataset was 98 minutes long. Both datasets contain real-world recorded DVL measurements. Using \textbf{EM5}, we showed that DCNet outperforms the baseline approach by achieving an 80\% improvement in calibration time. That is, instead of 100 seconds required for the calibration procedure, ours require only 20 seconds. Moreover, we demonstrate that during those 20 seconds the AUV should maintain a nearly constant velocity trajectory and does not require any complex maneuvering. In terms of the velocity accuracy, \textbf{EM5} obtained an average of 70\% improvement over the baseline in terms of the velocity RMSE. In summary, this paper presents an end-to-end data-driven calibration approach that reduces the calibration process complexity while maintaining high accuracy. This holds true only for low performance (low-cost) DVL. Therefore, our work opens up the possibility of using low-cost DVLs and achieving higher accuracy. As a consequence, different types of marine robotic systems will be able to use low-cost DVLs with high accuracy opening them up for use in new applications. 

\section*{Acknowledgement}
\noindent Z.Y. is supported by the Maurice Hatter Foundation and
University of Haifa presidential scholarship for outstanding
students on a direct Ph.D. track.



\bibliographystyle{IEEEtran}
\bibliography{bio.bib}

\end{document}